# Using Belief Functions for Uncertainty Management and Knowledge Acquisition: An Expert System Application

Dr. Mary Deutsch-McLeish
Paulyn Yao
Department of Computing and Information Science
Dr. Tatiana Stirtzinger (DVM, PhD)
Department of Pathology
University of Guelph
Guelph, Ontario, Canada N1G 2W1

This paper describes recent work on an ongoing project in medical diagnosis at the University of Guelph. A domain on which experts are not very good at pinpointing a single disease outcome is explored. On-line medical data is available over a relatively short period of time. Belief Functions (Dempster-Shafer theory) are first extracted from data and then modified with expert opinions. Several methods for doing this are compared and results show that one formulation statistically outperforms the others, including a method suggested by Shafer. Expert opinions and statistically derived information about dependencies among symptoms are also compared. The benefits of using uncertainty management techniques as methods for knowledge acquisition from data are discussed.

## 1. INTRODUCTION

Hospital database systems have been primarily used for billing and admissions procedures. More recently systems are being installed which collect on-line the results of many medical tests. A study of this data collected over time should prove valuable for the enhancement of medical expert systems. Although many epidemiological studies are carried out on medical data, often the data used does not represent the tests carried out a particular hospital and upon which immediate diagnoses are made. The data collected by these systems is dependent on the type and number of cases coming to that particular hospital. This can result in small sample sized for certain problems and possibly no information in some cases. Finding the best methods to analyze this type of data and designing a system which can incorporate expert opinions is the focus of a current project at the University of Guelph.

The Ontario College of Veterinary Medicine has an information system which stores a considerable amount of medical information on each patient including bacteriology, clinical pathology, parasitology, radiology and patient information such as age, sex, breed, presenting complaint, treatment procedures, diagnosis and outcome. In clinical pathology, much of the data is electronically generated by the lab equipment. The system is run using the DBMS ORACLE on a SEQUENT symmetry machine.

Earlier work involved the diagnosis of surgical cases in equine abdominal problems. Here several methods were compared, such as Bayesian Inductive Inference [Self and Cheeseman,1987], the PLS1 algorithm [Rendell, 1986], ID3 [Quinlan, 1983]. Statistical methods such as discriminant analysis and logistic regression [Mathews, 1985] were compared with the machine learning methods. However it was found that a method using weights of evidence [Good, 1964] did as well or better than any of the other methods. This method (similar to Bayesian techniques) had the additional advantage that expert opinions could be used to modify data-derived values. Significant symptom groups were found and diagnosis was based on a relatively large number of symptoms and not just a few. This was an advantage in the presence of missing data at the diagnostic stage. Results of these earlier studies can be found in the papers by [McLeish, Cecile et al., 1988, 1990].

The features exhibited by this latter methodology were sought for another domain. This new domain concerns the pre-biopsy diagnosis of liver diseases in small animals, in particular canine liver disease. Clinicians can accurately tell 75% of the time whether or not liver disease is present. They can only predict about 15% of the time the specific type of liver disease. Laboratory data in conjunction with expert opinions is being used to more accurately predict the specific type of problem. The cost of doing laboratory tests is about 20 times cheaper than the cost of performing a biopsy, which carries with it risks due to anesthesia, hemorrhage, infections and poor healing. A study [Myers, 1986] shows how laboratory information can be used, but it is based on expert opinions rather than a study of data.

On the liver disease domain the amount of data available was small compared to the number of outcomes and symptoms. Considering also the expert's inability to predict individual outcomes, it was decided to extract Belief Functions (Shafer) from data. Belief functions are strengths assigned to subsets of an outcome space, rather than individual diseases. These were then modified with expert opinions. (This was carried out after examining the values found from data, unlike the reverse procedure used in the probabilistic system in [Spiegelhalter, Franklin and Bull, 1989]).

continuous variables, eg. ablumins, calcuim, glucose, cholesterol, total protein, red blood cell, lymphocytes, reticulocytes, etc.). Initially we had only 150 cases and solicited expert opinions to reduce the outcome space to set of 14 classes based on the similarity of the lesions and the pathogenesis of liver diseases. Examples of these classes are Primary and Metastic Tumors, Hepatocellular Necrosis, Hepatic Fibrosis and Cirrhosis, Hepatic Congestion, Hepatic Atrophy and Hypoplasia etc. Even on these 14 classes, the data sample is not large enough to be convincingly discriminatory. As our intention was to solicit expert opinions to modify our statistically determined results and these were also not always well defined on singleton classes, it was decided to use the belief function approach from Dempster-Shafer theory. (In the equine domain, subsets of the small outcome space were meaningless and substantially more data per outcome was available). The problem formulation does not lend itself to the hierarchical or network approaches of [Zarley et al., 1988] for example. Although tests of correlations between parameters were carried out, detailed hierarchies were not well known.

In D-S theory, the frame of discernment $\theta$ is not the set over which a probability measure is defined. The power set, $2^\theta$, is the basic set upon which judgments are made. The subtle distinction between a probability mass assigned to a singleton and a larger subset containing that singleton is not immediately arrived at by considering frequency of occurrence information. If different outcomes were very well discriminated over some range of a continuous variable (ie, normal range for some medical symptom), values could be easily assigned. Unfortunately plots of our 40 symptoms in the liver disease domain revealed this not to be the case (even with outcomes considered as members of $2^\theta$). The range of continuous variables to be considered as individual symptoms were specified by the experts. These regions could be further refined to produce a greater degree of discrimination, but the computational complexity of the domain was already so high that we decided to keep to the expert defined ranges.

Method 1

There are several ideas which present themselves of how to make use of data to assign mass and belief functions in such situations. One idea comes from [14], where a support function $S_x(A)$ is defined by the following or all non-empty A:
$$S_x(A) = 1 - Pl_x(\overline{A}),$$
where the plausibility function
$$Pl_x(A) = \frac{(\max q_{\theta_i}(x), \theta_i \in A)}{(\max q_{\theta_i}(x), \theta_i \in \theta)}$$

and $\overline{A}$ stands for A complement.

Here $\theta_i$ is an outcome and x is a symptom. The functions $\{q_{\theta_i}\}_{\theta_i \in \theta}$ is a statistical specification which obeys the rule that x renders $\theta_i \in \theta$ more plausible than $\theta_i'$ whenever $q_{\theta_i}(x) > q_{\theta_i'}(x)$. A possible choice for $\{q_{\theta_i}\}$ are likelihood functions, see [Shafer].

One may work out what the associated mass functions are. (This is necessary for the method of implementation).

**Theorem 2.1**

If $f_x(\theta_i)$ represents the frequency of occurrence of outcome $\theta_i$ given symptom x and these outcomes are sorted in descending order by the f function for each x, then the non-zero mass functions are given by the formulae:
$$m_x(\theta_1) = \frac{f_x(\theta_1) - f_x(\theta_2)}{f_x(\theta_1)}$$
$$m_x(\theta_1 \ldots \theta_j) = \frac{f_x(\theta_j) - f_x(\theta_{j+1})}{f_x(\theta_1)}$$
$$m_x(\theta_1 \ldots \theta_n) = \frac{f_x(\theta_n)}{f_x(\theta_1)}$$

where $\theta = (\theta_1 \ldots \theta_n)$ and $f_x(\theta_i) \geq f_x(\theta_j)$, $i \leq j$.

**Proof** The consonance of the belief function resulting from the formula for $S_x(A)$ is noted in Theorem 11.1 in [Shafer, 1976]. Thus its focal elements can be arranged in order so that each one can be contained in the following one. It is not difficult to determine that the foci are the sets given above. To determine the belief values consider $\theta_i$
$$Bel\{\theta_1\} = 1 - \frac{f_x(\theta_2)}{f_x(\theta_1)} = m_x(\theta_1)$$
by definition of $S_x(A)$

Consider now
$$Bel\{\theta_1, \theta_2\} = 1 - \frac{f_x(\theta_3)}{f_x(\theta_1)} - Bel\{\theta_1\}$$
Therefore $m_x\{\theta_1, \theta_2\} \neq 0$ and equals
$$Bel\{\theta_1, \theta_2\} - m_x(\theta_1) = \frac{f_x(\theta_2) - f_x(\theta_3)}{f_x(\theta_1)}$$
In general
$$Bel\{\theta_1 \ldots \theta_j\} = 1 - \frac{f_x(\theta_{j+1})}{f_x(\theta_1)}$$
and
$$m\{\theta_1 \ldots \theta_j\} = Bel\{\theta_1 \ldots \theta_j\} - Bel\{\theta_1 \ldots \theta_{j+1}\}$$
(due to consonance)
$$= \frac{f_x(\theta_j) - f_x(\theta_{j+1})}{f_x(\theta_1)}$$

Now support $(\emptyset) = 0$ and under the assumption that the m values sum to 1,
$$m_x(\theta) = \frac{f_x(\theta_n)}{f_x(\theta_1)}$$

The question of extracting suitable mass functions from data is mentioned in [Shafer, 1976] and a suggestion is made. However, it is stated that the ideas given "*are not implied by the general theory of evidence exposited in the preceding chapters. Rather, these assumptions must be regarded as conventions for establishing degrees of support, conventions that can be justified only by their general intuitive appeal and by their success in dealing with particular examples.*" In Section 2 of this paper, this method (converted to a mass function instead of a plausibility function) is explained. Several other methods are outlined and algorithms provided for their execution.

Section 3 explains the modifications of symptom sets by expert opinions and by statistical analysis. Expert opinions were sought both before and after seeing the results of studying the data. Different methods for modifying the data-derived belief functions with expert opinions are also presented in this section.

Section 4 summarizes a few significant results. Very extensive testing with new data was actually carried out. There were may permutations of methods and data sets. Belief functions, mass functions and belief intervals were computed and an intricate comparison process was derived for reporting system performance against true singleton diagnoses. This testing showed how one of our own algorithms for finding belief functions out performed the method suggested by Shafer. Using statistical data generally out performs experts in choosing related symptoms. One method of incorporating expert opinions into the data derived belief functions is shown to enhance performance.

Section 5 comments on the advantages gained by using such a method on a domain of this type and reports on some continuing research.

### 1.1 A REVIEW OF BASIC TERMS FROM DEMPSTER-SHAFER THEORY

In Dempster-Shafer theory, all the possible finite set of hypotheses are represented as $\theta$, a frame of discernment. The hypothesis in $\theta$ are assume to be mutually exclusive and exhaustive. Each proper subset of $\theta$ is called a *proposition*. Therefore, for a given $\theta$ of size $n$, there will be $2^n$ possible propositions which one can make.

The function $m : 2^\theta \rightarrow [0,1]$ is called a *basic probability assignment*, whenever
1. $m(\emptyset) = 0$ and 2. $\sum_{Y \subseteq \theta} m(Y) = 1$

The first condition means that there is no belief committed to an empty set. The second condition means that the total belief that one commits has to equal 1.

If we have $m(A) = s$, where $A \subseteq \theta$, then the remaining probability mass is assigned to $\theta$, ie, $m(\theta) = 1 - s$. Thus, $m(A) + m(\theta) = 1$. The quantity $m(A)$ is a measure of belief committed *exactly to A*, where $A \subseteq \theta$, whereas the quantity $m(\theta)$ is a measure of belief that *remains unassigned* after commitment of belief to various proper subset of $\theta$. Generally, the quantity $m(X)$, where $X \subseteq \theta$, is called a *basic probability number* (BPN), which means a degree of strength in favor of proposition $X$. Sometimes, it is also called *probability mass*. Furthermore, if $m(A) > 0$, where $A \subseteq \theta$, the subset $A$ is called a *focal element* of a belief function $Bel(A)$ over $\theta$.

The function $Bel : 2^\theta \rightarrow [0,1]$ is called a *belief function* whenever:
1. $Bel(\emptyset) = 0$  2. $Bel(\theta) = 1$  and
3. $Bel(A) = \sum_{B \subseteq A} m(B)$, $\forall A \subseteq \theta$

The plausibility function which is denoted as $Pl(A)$ expresses the extent to which one finds that the set $A$ is credible or plausible, or it expresses the extent to which one finds that the set $A$ is doubtful, where $A \subseteq \theta$. The $Pl(A)$ can be represented in terms of a belief function as
$$Pl(A) = 1 - Bel(\overline{A}).$$
and in turn, $Bel(A)$ can also be represented in terms of $Pl(A)$ as
$$Bel(A) = 1 - Pl(\overline{A}), A \subseteq \theta$$
A belief function is said to be *consonant*, if its focal elements are nested as $A_1 \subseteq A_2 \ldots A_n$. Let $A$ be a set $A_i$, and $B$ be a set $A_j$ where $i \neq j$.

### 2. METHODOLOGY

A number of considerations motivated the development of an automated system for this domain.
(1) The relevant data is available in the database system at the hospital and is generated on-line as results are produced by the biochemistry analyzers.
(2) The expert's success rate has not been particularly good in this domain, partly due to the large number of input parameters and outcome variables involved. Papers on the use of laboratory data to identify liver diseases [Goldberg 1987, Hall, 1985] state that no single test is superior for identifying and detecting liver diseases. Thus it is desireable to try to analyze and interpret many parameters.
(3) The demands on the few specialists often leaves interns and other students with considerable responsibilities. An aid to them both as a learning tool and a diagnostic aid would be very valuable.

The interconnections between the variables in this domain are not well understood and attempts at producing a network of causal relations resulted in a very complicated structure of doubtful accuracy. The outcome space consists of 30 different types of liver disease and the laboratory tests produced 19 biochemical parameters and 21 hematologic parameters (all This result was the form used in the implementation, which was written to accommodate general mass function assignments (not necessarily support functions). As indicated in the introduction, however, there is no real theoretical justification for this method and Shafer himself suggests experimentation to determine an appropriate method for a particular application. Some other methods are presented below.

### Method 2

A search for a simple support function for each symptom was carried out by looking for individual symptoms or sets of symptoms with likelihoods greater than 0.5. As the values were based on frequency data, only one singleton could have such a high value. If none have a suitably high value, then sets of size two are examined and the pair with the highest value greater than 0.5 is chosen. The process is continued to higher sized sets if necessary.

One problem concerns what to do with the remaining mass. Although D-S theory normally would put it on $\theta$, the remaining frequencies of occurrences are on the complement of the support subsets. Two methods were actually tried on the data: 2A: putting the remaining mass on A where $m(A) \geq 0.05$ and 2B: assigning it to $\theta$. In actually implementing this, there are problems with thies etc. The following algorithm describes the precise method.

An algorithm for main part of method 2A (2B is similar):
- sort all the freq(A), where $A \subseteq \theta$ and A is a singleton, from the highest to the lowest values.
- from the highest to the lowest values, do the following:
  - if freq(A) > 0.5 then
    * $B \leftarrow A$
    * $m(B) \leftarrow freq(A)$
  - else
    while $m(B) \leq 0.5$
    {to get total m(B) > 0.5}
    * $B \leftarrow B \cup A$
    * $m(B) \leftarrow m(B) + freq(A)$
    * get the next freq(A)
  - get the next lower freq(A), say freq(X) - if there is any
  - while freq(X) equal to freq(A)
    {to include all the singletons in which their frequencies are the same}
    * $B \leftarrow B \cup X$
    * $m(B) \leftarrow m(B) + freq(X)$
    * get the lower freq(X), and assign it to freq(X)
  - while freq(X) > 0
    {collect all singletons in which their proportions are greater than zero}
    * $C \leftarrow C \cup X$
    * $m(C) \leftarrow m(C) + freq(X)$
    * get the next lower freq(X), and assign it to freq(X)
- where $B \subseteq \theta$, $A \subseteq \theta$, $X \subseteq \theta$, A, X are singletons.

### Method 3

Several variations on a computationally intense method were considered. These assigned mass function values based on observations of frequencies to all subsets and then used a variety of normalization techniques: normalizing over the entire $2^\theta$ or normalizing across subset groups of the same size first and then normalizing at the end. Variations were also produced in which $\theta$ was assigned values of $\emptyset$ or 1 before normalizing. The normalization had the affect of spreading the mass over singleton and larger subsets in such a way that the relative frequencies are preserved. However, there is not really any justification for the amount of weight that is ultimately assigned to pairs etc. instead of singletons. In the absence of data sets that are well discriminated within a symptom region, this method essentially spreads some of the mass, that would have been assigned to singletons in a probability model, out over higher order subsets in a uniform manner. This then always allows for some doubt as to whether the outcome is exactly a singleton.

## 3. INCORPROATION OF EXPERT OPINIONS

### 3.1 MODIFICATION OF THE BASIC PROBABILITY MASS ASSIGNMENTS

An expert was asked to provide BPA's for all the laboratory parameters. A description of how this was done and some examples follow. Two methods of modifying the BPA's were used: *part-* or *all-* modifications.

In the *part*-modification, the generated BPA's of a *laboratory parameter class* are modified into the expert-defined BPA's of that class, if and only if the expert-defined BPA's of such class are known propositions, ie. they are not $m(\theta) = 1.0$ or non-ignorance propositions. Thus, if the expert-defined BPA's of a parameter class are unknown, ie.
$\{\theta\} = 1.0$, there is no modification on the generated BPA's of that class. For example, the expert's opinion on AALB are:

| AALB | | |
|---|---|---|
| above reference intervals | $m(\theta) = 1.0$ | |
| within reference intervals | $m(\theta) = 1.0$ | |
| below reference intervals | $m(\{6\}) = 0.46$ | |
| | $m(\{9\}) = 0.27$ | |
| | $m(\{13\}) = 0.27$ | |

and the original BPA's for AALB are:

In the second methodology an expert was asked to remove some of the laboratory parameters when they thought there were some correlations among the parameters. This set of parameters was called 'expert 1' decision. Lastly, the correlation graphs were shown to the same expert and the expert was again asked to remove parameters. The results from all three of these processes were different and were tested separately. For example, for the biochemical parameters, the parameters ACA, ACL, ACBILI, ACREAT, AP and ANA were removed in the second methodology and AAMY, ACHOL, ACA, ACREAT, ACPK, AAGT, AP, AK and ANA were removed after the last method. These last two methods will be referred to as expert 1 and expert 2 decisions.

## 4. SUMMARY OF RESULTS

Some preliminary results based on an initial data set as described earlier are published in [McLeish, Yao, Stirtzinger and Cecile, 1989]. The expert opinion modifications were not fully incorporated at that time. In this study, 30 different types of liver diseases are represented by 241 cases. The thirty groups are reclassified into 14 groups based on the similarity of the lesions and the pathogenesis of the liver disease. Forty new cases were used as test cases; the 241 cases were used as training cases. Before correlations were incorporated, 40 laboratory variables were used, further split into regions or normal, high and low readings. The result from each run was compared with the result from either biopsy or necropsy. The result from a run is called *observed outcome set*; while the result from either biopsy or necropsy is called *expected outcome set*. Each comparison between the observed and expected outcome sets was categorized into one of the following:
1. Precise-Match (PM)
   where the singleton of the observed outcome is the same as in the singleton expected outcome.
2. Imprecise-Match (IM)
   where the singleton expected outcome exists in the non-singleton set of the observed outcomes.
3. Non-Match (NM)
   when no element of the observed outcome sets exists in the expected outcome set.

Later, the percentage of each category is calculated in terms of each methodology.

With respect to the above categories, we prefer each method to have a high percentage of Precise-Match and a low percentage of Non-Match. The Imprecise-Match can be in any range as long as it does not contribute to the *non-match* category.

### 4.1 METHODS IN WHICH THE BPA'S ARE ACQUIRED FROM DATA

Table 4.1  Percentage Result Based on Methods of Acquiring BPA's from Data

| Category | Method 1 | Method 2 | Method 3 |
|---|---|---|---|
| PM | 42.5 | 65.0 | 35.0 |
| IM | 20.0 | 7.5 | 2.5 |
| NM | 37.5 | 27.5 | 62.5 |

The above table shows that method 2 - the method which assigns the remaining BPN's to the remaining disease groups which have their singleton proportions greater than 0 - provides a higher performance than method 1 - the method which assigns the remaining BPN's to $\theta$. This happens because method 2 has a smaller number of liver disease groups in the second set of a given probability mass than in method 1, which always has the total distinct number of liver disease groups. Thus, in method 2, when all the evidences are combined, the common liver disease group will get its evidence supported more than that in method 1. Furthermore, method 2 provides a high percentage of PM which were prefered in determining the effectiveness of different methods. Method 2 statistically outperformed methods 1 and 3 ($\alpha$ level, 0.05).

Method 3, which is the approach suggested by Shafer, turns out to have the worst performance. The percentage of NM is about 25 to 35% higher than method 1 and 2 respectively. Since method 3 has the worst performance, the BPA's which is generated by method 3 are not modified by expert's opinions.

### 4.2 USE OF EXPERT'S OPINIONS

Table 4.2  Percentage Result Based on Approaches of Acquiring BPA's from Data and an Expert

| Category | Method 6 | Method 7 | Method 8 | Method 9 |
|---|---|---|---|---|
| PM | 60.7 | 30.0 | 25.0 | 28.6 |
| IM | 10.7 | 2.5 | 7.5 | 10.7 |
| NM | 28.6 | 67.5 | 67.5 | 60.7 |

Here methods 6 and 8 represent part-modification of methods 2 and 1 respectively by expert opinions and methods 7 and 9 represent all-modifications. One can see that method 6 provides the highest percentage of PM and lowest percentage of NM. Thus, when BPA's are acquired from both data and expert, it is better to use *some* of the expert's opinions to modify the generated BPA's from Method 2.

In the following table, method 5 represents the case where all the BPA's were acquired from the expert alone.

AALB    above reference intervals    $m(\{5,6\}) = 0.6$  
                                         $m(\theta) = 0.4$  
         within reference intervals    $m(\{3\}) = 0.7$  
                                         $m(\theta) = 0.3$  
         below reference intervals    $m(\{2,1\}) = 0.8$  
                                         $m(\theta) = 0.2$

When the *part*-modification is applied, the generated BPA's for AALB are changed into:

AALB    above reference intervals    $m(\{5,6\}) = 0.6$  
                                         $m(\theta) = 0.4$  
         within reference intervals    $m(\{3\}) = 0.7$  
                                         $m(\theta) = 0.3$  
         below reference intervals    $m(\{6\}) = 0.46$  
                                         $m(\{9\}) = 0.27$  
                                         $m(\{13\}) = 0.27$

In *all*-modification, the generated BPA's of a *laboratory parameter* are modified into the expert-defined BPA's of that parameter, if and only if at least one class of that parameter has the expert-defined BPA's of known propositions. Thus, the above original BPA's of AALB are modified into:

AALB    above reference intervals    $m(\theta) = 1.0$  
         within reference intervals    $m(\theta) = 1.0$  
         below reference intervals    $m(\{6\}) = 0.46$  
                                         $m(\{9\}) = 0.27$  
                                         $m(\{13\}) = 0.27$

Therefore, the *part*-modification is the modification which lies on the *laboratory parameter class*, and the *all*-modification is the modification which lies on the *laboratory parameter* level. Both types of modifications do not attempt to change the laboratory parameters in which all the classes have the unknown expert-defined BPA's.

### 3.2 MODIFICATION OF THE SET OF LABORATORY PARAMETERS

The input data was modified by removing one or more correlated laboratory parameters. This was carried out in three ways: (1) using statistical correlations (2) using expert opinions (3) using statistical correlations and expert opinions. In the first methodology, Pearson Correlation is used to indicate a linear correlation between two laboratory parameters from within the groups of biochemical and hematologic parameters. (Correlating parameters from biochemical and hematologic groups is not useful, since they are obtained from two different compartments of the peripheral blood. The biochemical data are obtained from the fluid phase, whereas the hematologic data are obtained from the cellular phase.) Correlated laboratory parameters which have the correlation coefficients either $\geq 0.5$ or $\leq -0.5$ are shown for the biochemical parameters in Table 3.2. The algorithm to decide which variables to remove is presented below:

- If there are 2 node in a graph,
  - an arbitrary node (laboratory parameter) is used
- else /* more than 2 nodes in a graph */
  - if all the nodes have the same number of branches
    /* 3 nodes in a graph */
    * keep the node with the larges sum of the Pearson correlation coefficients
    * remove the rest of the nodes
  - else /* more than 3 nodes in a graph */
    * the node with the most branches is kept, lets call this, node A.
    * while there is node which has not been visited
      /* till all the node is visited */
      · if there is a node X which is connected to the node A and to another node, say Y, node X is kept too.
    * remove the rest of the nodes

After applying this algorithm, one of the possible results is to remove the following laboratory parameters:
- AALB, AALT, ACBILI, ACPK, AAGT, AP, AGLUC, ATPROT, and AUREA from biochemical parameters

A similar process was carried out for the hematologic parameters.

Table 3.2

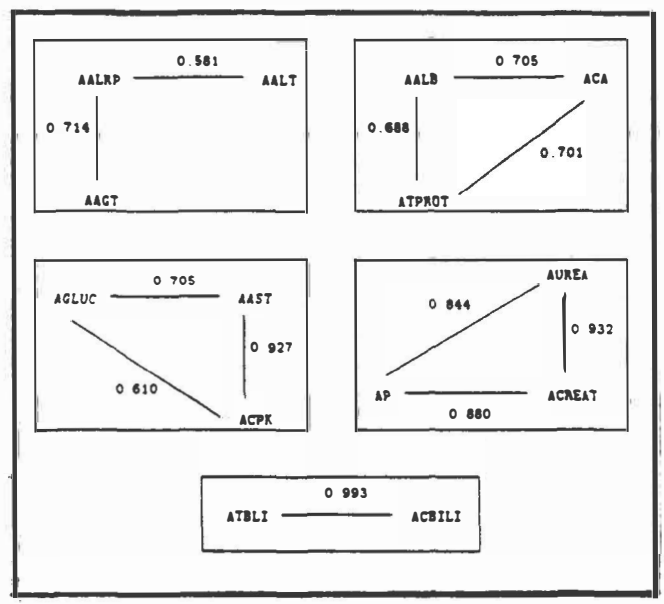

Table 4.3    Percentage Result Based on Types of Acquiring BPA's

| Category | Method 2 | Method 5 | Method 6 |
|---|---|---|---|
| PM | 65.0 | 9.1 | 60.7 |
| IM | 7.5 | 50.0 | 10.7 |
| NM | 27.5 | 40.9 | 28.6 |

2 and 6 are not significantly different. Thus, the ideal types of BPA's acquisitions are from data, and/or from data with *some* expert's opinions added to it.

Table 4.4    Average Percentage Result Based on Removal and Non-Removal Some of CorrelatedVariables

| Category | Non-Removal | Removal expert 1 | expert 2 | non-expert |
|---|---|---|---|---|
| PM | 47.5 | 48.3 | 40.8 | 47.5 |
| IM | 10.0 | 15.0 | 35.8 | 19.2 |
| NM | 42.5 | 36.7 | 23.4 | 33.3 |

This table shows the average of removal and non-removal some of the correlated variables. Even though there is no significant difference among expert 1, expert 2, and non-expert, the results show that the performance of the removal some of the correlated variables is improving about 5% to 19% over that of non-removal. Furthermore, it is difficult to say which kind of removal is the better than the other, since there was no significant difference among them. The one which provides the highest percentage of PM also provides the highest percentage of NM is expert 1. The one which provides the lowest percentage of NM also provides the lowest percentage of PM is expert 2. However, we still can conclude that the performance is improving when some of the correlated parameters are removed.

## 5.    CONCLUSIONS AND FURTHER WORK

Work by [Leaper et al, 1972] compares the performance of a diagnostic system using survey data, the unaided clinician, and clinician's estimates in diagnosing abdominal pain. They suggest the use of data in diagnostic systems, noting a well-known fact that "clinicians cannot analyze cases in a probabilistic sense, since they often have little idea of what 'true' probabilities are".

In our diagnostic system, we conclude that the performance when using past data with the removal of some of the correlated parameters is more effective than the performance when using expert's opinions only on the probability mass. When comparing the ability of the clinicians and the diagnostic system, the performance of the diagnostic system is about 60%-65% more effective than the performance of the clinicians in determining specific types of lesions. Norusis, [1975] and Szolovits, [1978] (see references) have stated that a large error rate may occur when a slight dependency exists. In our study, we have shown that if some dependent laboratory parameters are removed, the performance is slightly improved. We have also found that methods proposed by ourselves performed significantly better than that suggested in [Shafer, 1976].

The system has been implemented using 'C' on a Sequent Symmetry Machine. The response time is usually less than 1 minute. The time complexity can usually be kept to $O(2^n)$, when n = number of laboratory parameters. (Note that the BPA's are not simple enough to employ the methods of [Barnett, 1981]). The program output provides upper and lower probability values for the 14 possible outcomes.

Future research continues with the methodology and the type of data used. Additional information such as history of illness, conditions of patients etc. is being added to the database of laboratory data. The methodology is being modified to find a probability distribution for the continuous variables over the power set of the outcome space. This is being done by breaking up the ranges of the continuous variables into regions in which the outcomes (or outcome subsets) are reasonably well discriminated. (Not every data item will be used as a piece of evidence. This would be computationally infeasable for a problem of our size. see [Strat, 1984, Lowrance, 1986].)

We conclude by noting several special features of this system and methodology:

1. It assists clinicians in diagnosing specific types of liver diseases without performing biopsies
2. It provides single as well as grouped diagnoses with lower and upper probability values
3. The results of the data analysis can be modified subjectively (unlike most classical statistical and machine learning methods), especially when sample sizes are small
4. Missing values only affect the accuracy of the BPA's. No further data is lost (as in regression analysis, for example) nor do the complications caused by missing data in many tree building algorithms arise
5. All the variables are used to make a diagnosis, rather than relying on a few (as in some statistical methods). This fact is important in the presence of missing data at the final diagnostic stage (not just in the test data)
6. The system runs in real time with an essentially immediate response time.

390

# REFERENCES

Barnett, J.A., "Computational Methods for a Mathematical Theory of Evidence", IJCAI-1981, pp. 868-875.

Cecile, M., McLeish, M., Pascoe, P., Taylor, W., "Induction and Uncertainty Management Techniques Applied to Veterinary Medical Diagnosis", Proc. Fourth AAAI Workshop on Uncertainty in AI, Aug. 88, pp. 38-48.

Dempster, A., "A Generalization of Bayesian Inference", Journal of the Royal Statistical Society, Series B, 1968, 30, pp. 325-339.

Goldberg, D.M., Brown, D., "Advances in the Application of Biochemical Tests to Diseases of the Liver and Biliary Tract": Their role in Diagnosis, Prognosis, and the Elucidation of Pathogenetic Mechanisms. Clinical Biochemisty, 20, 1987, pp. 127-148.

Good, I.J., "Weight of Evidence, Corroboration, Explanatory Power, Information and the Utility of Experiments, JRSS B, 22, 1964, pp. 319-331.

Hall, R.L., "Laboratory Evaluation of Liver Disease", Symposium on Liver Diseases, 15, 1985, pp. 3-19.

Leaper, D.J., Horrocks, J., Staniland, J., DeDombal, F., "Computer-Assisted Diagnosis of Abdominal Pain Using 'Estimates' provided by Clinicians", British Medical Journal, 4, 1972, pp. 350-354.

Lowrance, J.D., Garvey, T., Strat, T., "A Framework for Evidential Reasoning Systems", AAAI, 1986.

McLeish, M., "Comparing knowledge Acquision and Classical Statistical Methods in the Development of a Veterinary Medical Expert System", Proc of The Interface in Statistics and Computing, Virginia, 1988, pp. 346-352.

McLeish, M., Cecile, M., Lopez-Suarez, A., "Database Issues for a Veterinary Medical Expert System", Fourth Int. Workshop on Statistical and Scientific Database Management, Rome, 1988, pp. 33-48.

McLeish, M., Cecile, M., "Enhancing Medical Expert Systems with Knowledge obtained from Statistical data", accepted for the Annals of Mathematics and Artificial Intelligence, Special Issue on AI and Statistics, to appear, 1990.

McLeish, M., Stirtzinger, T., Yao, P., Cecile, M., "Using Weights of Evidence and Belief Functions in Medical Diagnosis", AAAI Stanford Spring Symposium Series, A.I.M., March 1990, pp. 132-136

McLeish, M. Yao, P., Cecile, M., Stirtzinger, T., "Experiments using Belief Functions and Weights of Evidence Incorporating Statistical Data and Expert Opinions", Uncertainty Management Workshop Proceedings, Windsor, 1989, pp. 253-265.

Mathews, D., Farwell, V., "Using and Understanding Medical Statistics", Kanger Press, 1985.

Myers, J., "The Computer as a Diagnostic Consultant with Emphasis on Use of Laboratory Data", Clin. Chem., 32, 1986, pp. 1714-1718.

Norusis, M., and Jacquez, M.J., "Symptom Non-Independence in Mathematical Models for Diagnosis", Comput. Biomed. Res., 1975, pp. 156-172.

Quinlan, V.R., "Learning Efficient Classification Procedures and their Application to Chess End Games", Machine Learning, An Artifical Intelligence Approach, Tioga, 1983, pp. 463-482.

Rendell, L.A., "A General Framework for Induction and a Study of Selective Induction", Machine Learning, Vol. 2, 1986, pp. 177-226.

Shafer, G., "A Mathematical Theory of Evidence", Princeton University Press, 1976.

Self, M. and Cheeseman, P., "Bayesian Predication for Arificial Intelligence", Proceedings of the Third AAAI Workshop on Uncertainty Management, 1987, pp. 61-69 (Also, AAAI, 1988 paper).

Spiegelhalter, D., Franklin, R., Bull, K., "Assessment, Criticism and Improvement of Imprecise Subjective Probabilities for a Medical Expert System", 5th Uncertainty Management Workshop, Windsor, 1989, pp. 335-343.

Strat, T.M., "Continuous Belief Functions for Evidential Reasoning", AAAI, 1984, pp. 308, 313.

Szolovitz, P., and Pauker, S.G., "Categorical and Probabilistic Reasoning in Medical Diagnosis", Artificial Intelligence, 11, 1978, pp. 115-144.

Zarley, D., Hais, Y-T., Shafer, G., "Evidential Reasoning using DELIEF", Proceedings of the American Association of Artificial Intelligence, 1988, pp. 205-209.